\documentclass[sigconf]{acmart}
\usepackage{algorithm}
\usepackage{algorithmic}
\usepackage{booktabs}
\usepackage{bm}
\usepackage{color, soul}
\usepackage{graphicx}
\usepackage{latexsym}
\usepackage{makecell}
\usepackage{multirow}
\usepackage{multicol}
\usepackage{subfigure}
\usepackage{threeparttable}
\usepackage{tabularx}
\usepackage{url}
\usepackage{xcolor}

\newcommand{\nop}[1]{}
\AtBeginDocument{%
  \providecommand\BibTeX{{%
    \normalfont B\kern-0.5em{\scshape i\kern-0.25em b}\kern-0.8em\TeX}}}

\copyrightyear{2021} 
\acmYear{2021} 
\setcopyright{acmcopyright}\acmConference[KDD '21]{Proceedings of the 27th ACM
SIGKDD Conference on Knowledge Discovery and Data Mining}{August 14--18,
2021}{Virtual Event, Singapore}
\acmBooktitle{Proceedings of the 27th ACM SIGKDD Conference on Knowledge
Discovery and Data Mining (KDD '21), August 14--18, 2021, Virtual Event, Singapore}
\acmPrice{15.00}
\acmDOI{10.1145/3447548.3467196}
\acmISBN{978-1-4503-8332-5/21/08}

\begin{document}
\fancyhead{}
\begin{sloppy}

\title{Reinforced Iterative Knowledge Distillation for Cross-Lingual Named Entity Recognition}

\author{Shining Liang$^{1,2,*}$,~Ming Gong$^{3}$,~Jian Pei$^{4}$,~Linjun Shou$^{3}$,~Wanli Zuo$^{1,2}$,~Xianglin Zuo$^{1,2\dagger}$,\\~Daxin Jiang$^{3\dagger}$}
\affiliation{%
	\institution{$^1$College of Computer Science and Technology, Jilin University}
	\institution{$^2$Key laboratory of Symbolic Computation and Knowledge
Engineering, Ministry of Education}
	\institution{$^3$STCA NLP Group, Microsoft}
	\institution{$^4$School of Computing Science, Simon Fraser University}
}
\email{{liangsn17,zuoxl17}@mails.jlu.edu.cn, {migon,lisho,djiang}@microsoft.com, jpei@cs.sfu.ca, zuowl@jlu.edu.cn}

\renewcommand{\shortauthors}{Shining and Ming, et al.}
\renewcommand{\thefootnote}{\fnsymbol{footnote}}

\begin{abstract}

\footnotetext[1]{Work done during the first author's internship at Microsoft STCA.}
\footnotetext[2]{Daxin Jiang and Xianglin Zuo are the corresponding authors.}

Named entity recognition (NER) is a fundamental component in many applications, such as Web Search and Voice Assistants. Although deep neural networks greatly improve the performance of NER, due to the requirement of large amounts of training data, deep neural networks can hardly scale out to many languages in an industry setting. To tackle this challenge, cross-lingual NER transfers knowledge from a rich-resource language to languages with low resources through pre-trained multilingual language models. Instead of using training data in target languages, cross-lingual NER has to rely on only training data in source languages, and optionally adds the translated training data derived from source languages. However, the existing cross-lingual NER methods do not make good use of rich unlabeled data in target languages, which is relatively easy to collect in industry applications.  To address the opportunities and challenges, in this paper we describe our novel practice in Microsoft to leverage such large amounts of unlabeled data in target languages in real production settings. To effectively extract weak supervision signals from the unlabeled data, we develop a novel approach based on the ideas of semi-supervised learning and reinforcement learning. The empirical study on three benchmark data sets verifies that our approach establishes the new state-of-the-art performance with clear edges. Now, the NER techniques reported in this paper are on their way to become a fundamental component for Web ranking, Entity Pane, Answers Triggering, and Question Answering in the Microsoft Bing search engine. Moreover, our techniques will also serve as part of the Spoken Language Understanding module for a commercial voice assistant. We plan to open source the code of the prototype framework after deployment. 

\end{abstract}




\maketitle

\section{Introduction}\label{sec:intro}

\emph{Named entity recognition (NER)}~\cite{nadeau2007survey} identifies text spans that belong to predefined entity categories, such as persons, locations, and organizations. For example, in the sentence ``John Doe wrote to the association of happy entrepreneurs.'', NER may identify that the first two words, ``John Doe'', refer to a person, and the last five words, ``the association of happy entrepreneurs'' refer to an organization.  As a fundamental component in Natural Language Processing (NLP), NER has numerous applications in various industrial products. For example, in a commercial Web search engine, such as Microsoft Bing, NER is crucial for Query Understanding~\cite{liu2019user}, Web Information Extraction~\cite{dong2020multi}, and Question Answering~\cite{Yuan2020EnhancingAB,Shou2020MiningIR}. For voice assistants such as Siri, Alexa, and Cortana, NER is a key building block for Spoken Language Understanding (SLU)~\cite{TUR2011}.  For global companies, such as Microsoft, cross-lingual NER is critical to deploy and maintain their products across hundreds of regions with a large number of languages (typically over one hundred).

Recently, deep neural networks achieve great performance in NER~\cite{liang2020bond,DBLP:conf/naacl/AkbikBV19}. However, deep neural network models typically require large amounts of training data, which presents a huge challenge for global companies to deploy and maintain their products across different regions with many languages. Importantly, labeling training data is not a one-off effort, instead, maintaining high-quality NER models requires periodical training data refresh, e.g., tens of thousands of new annotated instances every few months per language. Moreover, with the evolving of products, there are often needs for schema update, e.g., adding more classes of named entities to be recognized, merging some existing classes, or retiring some classes. Such schema updates cause extra cost in adjusting or even relabeling training data to comply with new schema. Although the crowd-sourcing approach 
can substantially reduce the cost of data labeling, when data refreshes, schema updates, as well as a large number of languages are considered, it is still too expensive, if not unrealistic at all, to manually label training data at an industrial scale. In addition to financial constraints, hiring crowd-sourcing workers, building labeling guidelines and pipelines, and controlling labeling quality especially on low resource languages are also challenging and time-consuming. Therefore, scaling out NER to a large number of languages remains a grand challenge to the industry.

To reduce the cost of human labeling training data, \emph{cross-lingual NER} tries to transfer knowledge from rich-resource (source) languages to low-resource (target) languages. This approach usually pre-trains a multilingual model to learn a unified representation of different languages (such as mBERT~\citep{DBLP:conf/naacl/DevlinCLT19}, Unicoder~\citep{huang2019unicoder}, and XLM-Roberta~\citep{DBLP:conf/acl/ConneauKGCWGGOZ20}). Then the pre-trained model is further fine-tuned using the training data in the source language, and is applied to other languages~\cite{DBLP:conf/aaai/BariJJ20,DBLP:conf/emnlp/WuD19}. Although this approach has shown good results for classification tasks, the performance on sequence labeling tasks, such as NER and SLU, is still far from perfect~\cite{li2020mtop,DBLP:conf/emnlp/WuD19}. Table~\ref{tab:dmt} compares the NER performance in English versus that in some target languages. Following~\cite{moon2019towards}, we fine-tune mBERT with English data and directly test on the target languages. A dramatic drop in F1 score in every target language clearly indicates a big performance loss.

\begin{table}[t]
\setlength{\abovecaptionskip}{3pt}
\setlength{\belowcaptionskip}{-3pt}
\centering
\small
\caption{The performance comparison between NER performance in English and some target languages.  Following~\cite{moon2019towards}, we fine-tune mBERT with English data and directly test on the target languages.
}
\resizebox{1.0\linewidth}{!}{
\begin{tabular}{l | c c c c } 
 \toprule
   Languages&  \textbf{English}  &\textbf{Spanish} &    \textbf{Dutch} &   \textbf{German} \\
 \midrule
    F1 Score & 90.87 & 75.56\  (-15.31) & 78.86\  (-12.01)& 71.94\  (-18.93)\\
 \bottomrule
\end{tabular}
}
\label{tab:dmt}
\vspace{-10pt}
\end{table}

To enhance the transferability of cross-lingual models, several methods convert training examples in a source language into examples in a target language through machine translation~\cite{DBLP:conf/emnlp/MayhewTR17,li2020mtop}. The annotation of entities is derived through word or phrase alignments between source and target languages~\cite{DBLP:conf/emnlp/XieYNSC18,DBLP:conf/acl/NiDF17}. Despite the improved transferability across languages, this approach still suffers from several critical limitations. First, parallel data and machine translators may not be available for all target languages. Second, translated data may not be diverse enough compared to real target data, and there may exist some translation artifacts in the data distribution~\cite{ARTETXE2020}. Finally, there are both translation errors and alignment errors in translated data, which hurt the performance of models~\cite{DBLP:conf/acl/NiDF17}.

In this paper, we describe a different approach to cross-lingual NER practiced in the Microsoft product team. Our approach is based on the industry reality that in real product settings, it is often feasible to collect large amounts of unlabeled data in target languages. For example, in both Web search engines and voice assistants, there are huge amounts of user queries or utterances recorded in the search/product logs. Compared with the existing approaches, our method does not need parallel data or machine translators. Moreover, the real user input is much larger in size and much richer in the diversity of expressions. Leveraging such rich and diversified unlabeled data is far from straightforward. Although some recent effort~\cite{DBLP:conf/acl/WuLKLH20} explore a semi-supervised knowledge distillation~\cite{hinton2015distilling} approach to allow a student model to learn the knowledge of NER from the teacher model through the distillation process, as shown in Table~\ref{tab:dmt} as well as the previous works~\cite{DBLP:conf/emnlp/WuD19,li2020mtop}, fine-tuning using English data alone often leads to inferior results for sequence labeling tasks.

\nop{
Recently, Wu~\textit{et al.} propose a semi-supervised approach, where unlabeled data in a target language is used as the media for knowledge transfer. The study applies a fine-tuned multilingual model $\mathcal{M}_0$ (the cased multilingual $BERT\textsubscript{base}$~\cite{DBLP:conf/naacl/DevlinCLT19}) to unlabeled data and uses the output of $\mathcal{M}_0$ as weak supervision to train a new model $\mathcal{M}_1$. This process is similar to the spirit of knowledge distillation, where $\mathcal{M}_0$ is a teacher model and $\mathcal{M}_1$ is a student model. Intuitively, However, 
}


In our approach, we adopt the knowledge distillation framework and use a weak model $\mathcal{M}_0$ tuned from English data alone as a starting point. The novelty of our approach is that we develop a reinforcement learning (RL) based framework, which trains a policy network to predict the utility of an unlabeled example to improve the student model $\mathcal{M}_1$. Then, based on the predicted utility,  the examples are selectively added to the knowledge distillation process. We observe that this screening process can effectively improve the performance of the derived student model. Moreover, we adopt a  bootstrapping approach and extend the knowledge distillation step into an iterative process: the student model derived from the last round can take the role of teacher model in the next round. With the guidance of the policy network, the noise in supervision signals, that is, the prediction errors made by teacher models is reduced step by step. The model evolves towards better performance for NER in each round, which in turn generates stronger supervision signals for the next round. 

We make the following contributions in this paper.  First, we target an underlying component in many industrial applications and call out the unaddressed challenges for cross-lingual NER. After analyzing various existing approaches to this problem and considering the industry practice, we propose to leverage large amounts of unlabeled data, which can often be easily collected in real applications. Second, we present our findings that by smartly selecting the unlabeled data in an iterated reinforcement learning framework, the model performance can be improved substantially. We develop an industry solution that can be used in many products built on NER.  Third, we conduct experiments on three widely used datasets and demonstrate the effectiveness of the proposed framework. We establish the new SOTA performance with a clear gain comparing to the existing strong baselines. Now, the NER techniques reported in this paper are on their way to become a fundamental component for Web ranking, Entity Pane, Answers Triggering, and Question Answering in the Microsoft Bing search engine. Moreover, our techniques will also serve as part of the Spoken Language Understanding module for a commercial voice assistant. 

The rest of the paper is organized as follows.  
We review the related work in Section~\ref{sec:related}, and present our method in Section~\ref{sec:method}. We report an empirical study in Section~\ref{sec:exp}, and conclude the paper in Section~\ref{sec:con}.
Table~\ref{tab:notations} summarizes some frequently used symbols.

\begin{table}[t]
\setlength{\abovecaptionskip}{3pt}
\setlength{\belowcaptionskip}{-3pt}
\centering
\small
\caption{Frequently used notations in the paper.}
\resizebox{1.0\linewidth}{!}{
\begin{tabular}{c|p{6.8cm}} 
 \toprule
 \textbf{Symbol} &    \textbf{Description}  \\
 \midrule
    $\tilde{\bm{p}}$& probability distribution of teacher/student model  \\
    $\mathcal{M}_{source}$ & model for source language\\
    $\mathcal{M}_{target}$ & model for target language  \\
    $\mathcal{M}_{k}$ & student model in the $k$-th distillation iteration \\
    $\pi$ & policy network used to select instances \\
    $\bm{S}_b^T$ & batch of state vectors\\
    $\mathcal{A}_b$ & sampled action with policy network \\
    $\mathcal{D}_{b}^{T}$,$\mathcal{D}_{b'}^{T}$ & batches of instances and selected instances in target language\\
\bottomrule
\end{tabular}}
\label{tab:notations}
\vspace{-10pt}
\end{table}

\section{Related Work}\label{sec:related}

Our approach is highly related to the existing work on cross-lingual NER, knowledge distillation, and reinforcement learning. In this section, we briefly review some most related studies, in addition to those discussed in Section~\ref{sec:intro}.

\emph{Zero-shot cross-lingual NER} seeks to extract entities in a target language but assumes only annotated data in a source language.
Pseudo training data in a target language may be generated by leveraging parallel corpus and word alignment models~\cite{DBLP:conf/acl/NiDF17} or by machine translation approaches~\cite{DBLP:conf/emnlp/MayhewTR17,DBLP:conf/emnlp/XieYNSC18}.
In addition to training using synthetic data, some approaches directly transfer models in source languages to target languages using a shared vector space to represent different languages~\cite{DBLP:conf/ijcnlp/WangPD17,DBLP:conf/iclr/YangSC17}.

Recently, pre-trained multilingual language models are adopted to address the challenge of cross-lingual transfer using only the labeled training data in the source language and directly transferring to target languages~\cite{DBLP:conf/emnlp/WuD19,DBLP:conf/emnlp/PfeifferVGR20,DBLP:conf/emnlp/WuD20}. Taking advantage of large-scale unsupervised pre-training, these methods achieve prominent results in cross-lingual NER. However, the performance in target languages is still unsatisfactory due to the lack of corresponding knowledge about target languages.
In this work, on top of those pre-trained multilingual models, we propose an iterative distillation framework under the guidance of reinforcement learning to enhance the cross-lingual transfer-ability using unlabeled data in target languages.


\emph{Knowledge distillation (KD)} is effective in transferring knowledge from a complex teacher model to a simple student model~\cite{hinton2015distilling,DBLP:conf/wsdm/YangSGLJ20}.
In a standard KD procedure, a teacher model is first obtained by training using golden standard labeled data. A student model is then optimized by learning from the ground-truth labels as well as mimicking the output distribution of the teacher model.
KD has also been used for cross-lingual transferring. For example,
Xu and Yang~\cite{DBLP:conf/acl/XuY17} leverage soft labels produced by a model in a rich-resource language to train a target language model on the parallel corpus.
Wu et al.~\cite{DBLP:conf/acl/WuLKLH20} train a teacher model based on a pre-trained multilingual language model and directly distill knowledge using unlabeled data in target languages.
Nevertheless, these methods directly perform knowledge distillation with all instances and do not address the subtlety that some samples may have a negative impact due to teacher model prediction errors.

In this work, we establish a reinforcement learning based framework to select unlabeled instances for knowledge distillation in cross-lingual knowledge transfer by removing the errors in teacher model predictions in the target language. This framework can be applied to not only NER, but also more cross-lingual Web applications, such as relation extraction and question answering.


\emph{Reinforcement learning (RL)}~\cite{sutton2018reinforcement} has been widely used in natural language processing, such as dialogue systems~\cite{DBLP:conf/aaai/Gonzalez-Garduno19} and machine translation~\cite{DBLP:conf/emnlp/WuTQLL18}.
Those methods leverage semantic information as rewards to train generative models.
Particularly, a series of studies use RL to select proper training instances.
For example, Wang et al.~\cite{wang2019minimax} leverage a selector to select source domain data closed to the target and accept the reward from the discriminator and the transfer learning module. 

Motivated by the above studies, in this work, we leverage RL to smartly select unlabeled instances for knowledge distillation. To the best of our knowledge, our work is the first to apply reinforcement learning for cross-lingual transfer learning.

\section{Methodology}\label{sec:method}

In this section, we first define the problem and review the preliminaries.  Then, we introduce our iterative knowledge distillation framework for cross-lingual NER.  Last, we develop our reinforced selective knowledge distillation technique.

\subsection{Problem Definition and Preliminaries}

We model cross-lingual named entity recognition as a sequence labeling problem. 
Given a sentence $\bm{x}={\{x_i\}}^L_{i=1}$ with $L$ tokens, a NER model produces a sequence of labels $\bm{y}={\{y_i\}}_{i=1}^L$, where $y_i$ indicates the category of the entity (or not an entity) of the corresponding token $x_i$. 
Denote by $\mathcal{D}_{train}^S= \{(\bm{x}, \bm{y})\}$ the annotated data in the source language, where the superscript $S$ indicates that this is a data set in the source language. In the target language, annotated data is not available for training except for a test set $\mathcal{D}_{test}^T$, where the superscript $T$ indicates that those are data sets in the target language. We also assume unlabeled data in the target language, denoted by 
$\mathcal{D}_{unlabeled}^T = \{\bm{x}\}$, which may be leveraged for knowledge distillation. Formally, zero-shot cross-lingual NER is to learn a model by leveraging both $\mathcal{D}_{train}^S$ and $\mathcal{D}_{unlabeled}^T$ to obtain good performance on $\mathcal{D}_{test}^T$.

An encoder $\mathcal{M}_{enc}$ is used to learn contextualized embedding and produce hidden states  $\bm{H}={\{\bm{h}_i\}}_{i=0}^L$, that is,
$
    \bm{H}=\mathcal{M}_{enc}(\bm{x}; \Theta)
$,
where $\Theta$ denotes the parameters of the encoder. 
Here we adopt two pre-trained multilingual language models, mBERT and XLM-Roberta (XLM-R) as the basic encoders separately, to verify the generalization of our method. In general, any encoding model that produces a hidden state $\bm{h}_i$ for the corresponding input token $x_i$ may be employed. 
For each token of the sequence, the probability of each category is learned by
$
    \bm{p}(x_i; \Theta) = \mathrm{softmax} (\bm{W} \cdot \bm{h}_i+\bm{b})
$, 
where $\bm{W} \in \mathcal{R}^{d_h \times c}$ and $\bm{b}\in\mathcal{R}^c$ are the weight and the bias term.

In general, the Knowledge Distillation (KD) approach~\cite{hinton2015distilling} uses the soft output (logits) of one large model or the ensemble of multiple large models as the knowledge and transfers the knowledge to a small/single student model. The distilled student model can achieve decent performance with high efficiency as well. Although KD was initially proposed for model compression, in this paper, we apply this approach to cross-lingual NER in order to transfer knowledge learned from the training data in the (rich-resource) source language to the (low-resource) target language. 

\begin{figure*}[t]
    \setlength{\abovecaptionskip}{3pt}
	\centering 
    \includegraphics[trim={0cm 0.6cm 0cm 0.9cm},clip,scale=0.35]{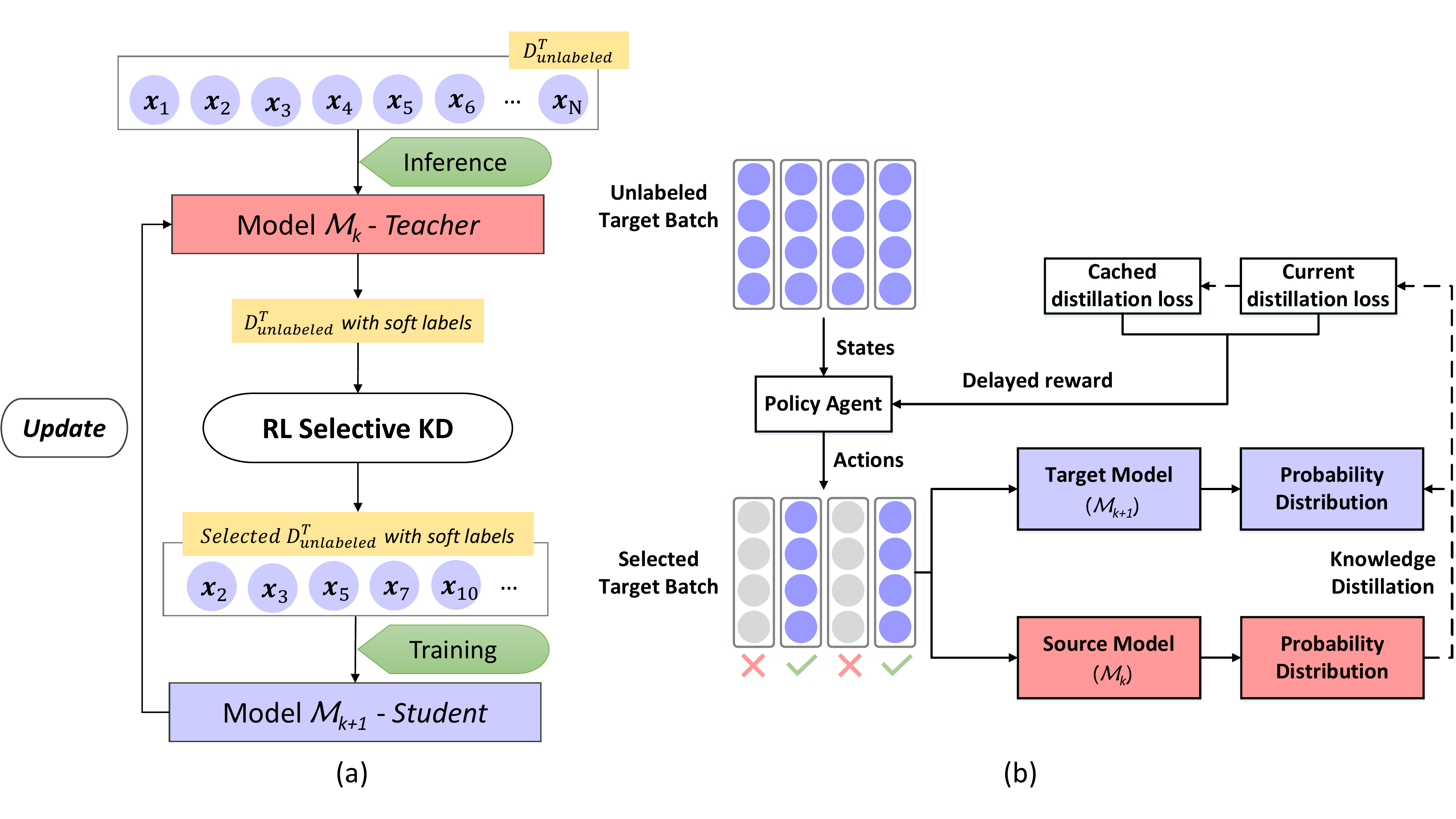}
	\caption{The architecture of our proposed method Reinforced Iterative Knowledge Distillation for cross-lingual NER. (a) The iterative KD framework. (b) RL based selective KD. Please note that model $\mathcal{M}_{0}$ is first obtained through fine-tuning the base model with the labeled data $\mathcal{D}_{train}^S$ in the source language. } 
	\label{fig:framework}
\vspace{-10pt}
\end{figure*}

For a NER task, given an unlabeled sentence $\bm{x} \in \mathcal{D}_{unlabeled}^T$, the distillation loss is the mean squared error loss between the predicted probability distributions of entity labels by the student model and that of the teacher model. To be specific, the loss with regard to $\bm{x}$ to train a student model is formalized as
$
        \mathcal{L}(\bm{x}; \Theta_{student}) =  \frac{1}{L}\sum_{i=1}^{L} \mathrm{MSE}(\tilde{\bm{p}}(x_i; \Theta_{teacher}), \; \tilde{\bm{p}}(x_i; \Theta_{student}))
$, 
where $\Theta_{teacher}$ and $\Theta_{student}$ are the parameters of the teacher model and the student model, respectively, $\tilde{\bm{p}}(\cdot; \Theta_{teacher})$ and $\tilde{\bm{p}}(\cdot; \Theta_{student})$ are the predicted label distributions of the teacher model and the student model, respectively, and $\mathrm{MSE}$ represents the mean squared error. The parameters of the teacher model are fixed during the training. In our knowledge distillation framework, both the teacher model and the student model share the same architecture (multilingual models) but with different parameter weights.

\subsection{Iterative KD for Cross-Lingual NER}\label{sec:IKD}

One challenge in cross-lingual NER is that the teacher model is trained by the source language but applied to the target language. Due to the differences between languages, the knowledge transferred from teacher model to student model may contain much noise. To address this challenge, we propose a framework {\em Reinforced Iterative Knowledge Distillation} (or RIKD for short).

The overall architecture of our method is shown in Figure~\ref{fig:framework}(a). A source multilingual model is first trained using the annotated data $\mathcal{D}_{train}^S$ in the source language.  The source multilingual model is leveraged as a teacher model to train a target model by transferring the shared knowledge from the source language to the target language. To reduce noise in knowledge transfer, we introduce a reinforced instance selector to select unlabeled data in the distillation step for better transfer learning.  Through the smart selection of examples in knowledge distillation, the student model can be improved over the teacher model on the target language. Therefore, we further iterate this  RL-based knowledge distillation step multiple rounds to drive the final target model, where the student model derived from the last round takes the role of teacher model in the next round.

In cross-lingual knowledge distillation for NER, although the source model $\mathcal{M}_0$ is only trained using the labeled data in the source language, it is capable of inferring directly on the cases in the target language, since it is benefited from the language-independent common feature space of pre-trained multilingual encoder and entity-specific knowledge of the labeled data. The cross-lingual transfer step aims to transfer language-agnostic knowledge from the source model to the model in the target language by minimizing the distance between the prediction distribution of the source model and that of the target model. 

\renewcommand{\algorithmicrequire}{\textbf{Input:}} 
\renewcommand{\algorithmicensure}{\textbf{Output:}}
\begin{algorithm}[t] \small
	\caption{: Reinforced Iterative Knowledge Distillation.} 
	\label{alg:ikd} 
	\begin{algorithmic}[1]
		\REQUIRE   \ 
		Iteration number $K$; Training steps number $J$; 
		Pre-trained model $\mathcal{M}_{0}$ in source language; Target language unlabeled data $\mathcal{D}_{unlabeled}^T$; Base NER model $\mathcal{M}_{NER}$ initialized using the pre-trained weights of mBERT or XLM-R.\\
		\ENSURE   \ 
		Distilled student model $\mathcal{M}_{K}$ for target language.\\
		\FOR{$k=1$ to $K$}
		\STATE Initialize a new model $\mathcal{M}_{k}$ with $\mathcal{M}_{NER}$.
		\FOR{$j=1$ to $J$}
		\STATE Sample a batch $\mathcal{D}_b^T$ from $\mathcal{D}_{unlabeled}^T$ then distill knowledge from $\mathcal{M}_{k-1}$ to $\mathcal{M}_{k}$ with $\mathcal{D}_b^T$ by Algorithm~\ref{alg:rl}.
		\ENDFOR
		\ENDFOR

	\end{algorithmic} 
\end{algorithm}

Specifically, given an instance $\bm{x}$ in the target language, we minimize the MSE of the output probability distributions between the source model and the target model, which is given by
$
        \mathcal{L}(\bm{x}; \Theta_{target}) =  \frac{1}{L}\sum_{i=1}^{L} \mathrm{MSE}(\tilde{\bm{p}}(x_i; \Theta_{source}), \; \tilde{\bm{p}}(x_i; \Theta_{target}))
$, 
where $\Theta_{source}$ and $\Theta_{target}$ are the parameters of the source model and the target model, respectively, and $\tilde{\bm{p}}(\cdot; \Theta_{source})$ and $\tilde{\bm{p}}(\cdot; \Theta_{target})$ are the predicted label distributions of the source model and the target model, respectively. 
This cross-lingual distillation step enables the target model to leverage the unlabeled data in the target language by mimicking the soft labels predicted by the source model to transfer knowledge from the source language. 

Inspired by the self-training paradigm~\cite{xie2020self, liu2020self}, where a model itself is used to generate labels from unlabeled data and the model is retrained using the same structure based on the generated labels, we further leverage the target model to produce the probability distributions of the training instances on the unlabeled data in the target language and conduct another knowledge distillation step to derive a new model in the target language. The training objective of this distillation step is formulated as
\begin{equation}\label{eq:distill} \small
    \mathcal{L}(\bm{x}; \Theta_{k}) =  \frac{1}{L}\sum_{i=1}^{L} \mathrm{MSE} (\tilde{\bm{p}}(x_i; \Theta_{k-1}), \; \tilde{\bm{p}}(x_i; \Theta_{k}))
\end{equation}
where $\Theta_{k-1}$ and $\Theta_{k}$ are the parameters of the model from the previous iteration and the new model to be trained, respectively, and $\tilde{\bm{p}}(\cdot; \Theta_{k-1})$ and $\tilde{\bm{p}}(\cdot; \Theta_{k})$ are the predicted label distributions of these two models, respectively. 

This iterative training step may be conducted multiple rounds by leveraging unlabeled data in the target language, which is relatively easy to obtain than labeled data.
In our experiments on the benchmark datasets, we find that three rounds can achieve decent and stable results. Algorithm~\ref{alg:ikd} shows the pseudo-code of our RIKD method.

\subsection{Reinforced Selective Knowledge Distillation}\label{sec:RKD}

Now let us explain the instance selector in our approach shown in Figure~\ref{fig:framework}(b). While conventional knowledge distillation directly transfers knowledge from the source model to the target model, the discrepancy between the source language and the target language may induce noise in the soft labels of the source model. As shown in Table~\ref{tab:dmt}, the model in the source language has low performances on other languages, thus the supervision from the predictions of the model in the source language may be noisy. 

To address this challenge, we use reinforcement learning to select the most informative training instances to strengthen transfer learning between the two languages (or two generations after the first round). We adopt this method in each round of RIKD. The major elements in our reinforcement learning procedure include states, actions, and awards.

\subsubsection{{State}} We model the state of a given unlabeled instance in the target language by a continuous real-valued vector $\bm{s}_i\in\mathcal{R}^{d_s}$. Given a target language instance $\bm{x}$,  we first obtain a pseudo labeled-sequence $\bm{\hat{y}}={\{\hat{y_i}\}}_{i=1}^L$ using the source model, and use the concatenation of a series of features to form the state vector. 

The first two features are based on the prediction results from the source model only. The \emph{number of predicted entities} indicates the informativeness of an instance by the source model, that is,
$
        f_{entities}=\sum_{i=1}^{L}(\hat{y_i} \ne O)
$, 
    where $O$ denotes non-entities in \emph{BIO} tagging schema following~\cite{DBLP:conf/emnlp/WuD19}.
The \emph{inference loss of the source model} indicates how confident the source model is about the prediction,
$
        f_{conf} =  -\frac{1}{L}\sum_{i=1}^{L} 
        log(\tilde{\bm{p}}(\hat{y_i}|\bm{x}; \Theta_{source}))
$.

The third feature is the \emph{MSE loss of the output probability distributions between the source model and the target model on the unlabeled instances}, which is
$
             f_{agree} =  \frac{1}{L}\sum_{i=1}^{L} \mathrm{MSE}(\tilde{\bm{p}}(x_i; \Theta_{source}), \; \tilde{\bm{p}}(x_i; \Theta_{target})) 
$. 
It combines the predictions by the source model and the target model. This feature is based on the intuition that the agreement between the source model and the target model may indicate how well the target model imitates the source model on the current instance. 

The fourth feature describes the internal representation as well as the output for the target model after seeing the example using the label-aware representation of the target model. 
        We convert the predicted label $\hat{y}_i$ into label embedding $\bm{u}_i$ through a shared trainable embedding matrix $\bm{U} \in \mathcal{R}^{c \times d_u}$, which is trained during the optimization process of policy network.
    Then a linear transformation is used to build a label-aware vector for each token:
$
        \bm{z}_i = \bm{W}_u([\bm{h}_i^t; \bm{u}_i])+\bm{b}_u 
$, 
    where $\bm{W}_u \in \mathcal{R}^{d_z \times (d_h +d_u)}$ and $\bm{b}_u \in \mathcal{R}^{d_z}$ are the weight matrix and the bias term, respectively, symbol $[;]$ denotes the concatenation operation and $\bm{h}_i^t$ is the hidden state of token $i$ at the last layer of the target model. A max-pooling operation over the length is used to generate the \emph{semantic feature}:
$
        f_{semantic}=\mathrm{MaxPooling}({\{\bm{z}_i\}}_{i=1}^L)
$. 
%

Last, we use the \emph{length of the target unlabeled instance} $f_{length} = L$, which relies on only the instance itself and is used to balance the effects of the instance length and number of predicted entities.
%
%
The parameters introduced to learn the state features are part of the policy network to be trained. 

\subsubsection{Action} We introduce a binary action space $a_i\in\{0,1\}$, which indicates whether to keep or drop the current instance $\bm{x}_i$ from a batch of instances to optimize the target model. A policy function $\pi(\bm{s}_i)$ takes the state as input and outputs the probability distribution where the action $a_i$ is derived from. The policy network $\pi$ is implemented using a two-layer fully connected network computed as
$
    \pi(\bm{s}_i) = \mathrm{sigmoid}(\bm{W}_2 \cdot \phi(\bm{W}_1 \cdot \bm{s}_i + \bm{b}_1) + \bm{b}_2)
$, 
where $\bm{W}_j$ and $\bm{b}_j$, respectively, are the weight matrix and the bias term of the $j$-th fully connected layer, and $\phi$ is the ReLU activation function.

\subsubsection{Reward} The selector takes as input a batch of states $\bm{S}_b^T$ corresponding to a batch of unlabeled data $\mathcal{D}_b^T$, and samples actions to keep a subset $\mathcal{D}_{b'}^T$ of the batch.
Since we sample a batch of actions in each updating step, the proposed method assigns one reward to a batch of actions.

We adopt delayed reward~\cite{wang2019minimax} to update the selector.
Recall that we aim to select a subset of informative training data to improve the cross-lingual NER task. For the $j$-th training step in iteration $k$, we sample actions and use the selected sub-batch to optimize the new target model $\mathcal{M}_{k}$ with parameters $\Theta_{k}^{j}$. The optimization objective is formulated as below according to Equation~\ref{eq:distill}.
\begin{equation} \small
\label{eq:rl-loss}
    \mathcal{L}(\mathcal{D}_{b'}^{T}; \Theta_{k}^{j}) =  \frac{1}{|\mathcal{D}_{b'}^{T}|}\sum_{\bm{x} \in \mathcal{D}_{b'}^{T}} \mathcal{L}(\bm{x}; \Theta_{k}^{j})
\end{equation}

Since we target the zero-shot setting where no labeled data is available in the target language including development set, we use the training loss delta on $\mathcal{D}_{b'}^{T}$ to obtain the delayed reward motivated by Yuan et al.~\cite{Yuan2020ReinforcedMS}, that is, 
\begin{equation} \small
\label{eq:reward}
    r = \mathcal{L}(\mathcal{D}_{b'}^{T}; \Theta_{k}^{j-1}) - \mathcal{L}(\mathcal{D}_{b'}^{T}; \Theta_{k}^{j})
\end{equation}
where $\mathcal{L}(\mathcal{D}_{b'}^{T}; \Theta_{k}^{j-1})$ is cached beforehand, and initialized by the training loss of the last warm-up training step when the reinforced training starts.

\subsubsection{Optimization} 
We use the policy-based RL method~\cite{DBLP:conf/nips/SuttonMSM99} to train the selector. Algorithm~\ref{alg:rl} shows the pseudo-code. 
First, we pre-train the target model without instance selection for several warm-up steps.
Second, for each batch $\mathcal{D}_{b}^{T}$, we sample actions based on the probabilities given by the policy network. Denote by $\mathcal{D}_{b'}^{T}$ the selected instances. 
Those selected target-language instances are then used as the inputs to perform knowledge distillation and update the parameters of the target model. The selector remains unchanged.
Last, we calculate the delayed reward $r$ according to Equation~\ref{eq:reward} and optimize the selector using this reward with cached states and actions, that is,
\begin{equation} \small
\label{eq:rl-opt}
    \mathcal{L}(\bm{S}_b^T,\Theta_{\pi}) = \frac{1}{|\bm{S}_b^T|}\sum_{i=1}^{|\bm{S}_b^T|}(-r)log [ (1-a_i)(1-\pi(\bm{s}_i))+ a_i\pi(\bm{s}_i) ]
\end{equation}

\renewcommand{\algorithmicrequire}{\textbf{Input:}} 
\renewcommand{\algorithmicensure}{\textbf{Output:}}
\begin{algorithm}[t]  \small
	\caption{: Algorithm for Reinforced Instance Selection.} 
	\label{alg:rl} 
	\begin{algorithmic}[1]
		\REQUIRE   \ 
		Policy network $\pi$; Target language unlabeled data $\mathcal{D}_{unlabeled}^T$; Teacher model $\mathcal{M}_{k-1}$ and student model $\mathcal{M}_{k}$ initialized using mBERT or XLM-R; Warm-up training steps $J_w$ and reinforced training steps $J_r$, $J = J_w + J_r$. \\
		\ENSURE   \ 
		Distilled student model $M_k$ in iteration $k$.
		\FOR{$j \in [1, J_w]$}
		\STATE Sample a random batch $\mathcal{D}_b^T$ from $\mathcal{D}_{unlabeled}^T$.
		\STATE Update the target model with $\mathcal{D}_{b}^{T}$ according to Equation \ref{eq:distill}.
		\ENDFOR
		
		\FOR{$j \in [1, J_r]$}
		\STATE Sample a random batch $\mathcal{D}_b^T$ from $\mathcal{D}_{unlabeled}^T$.
		\STATE Obtain states $\bm{S}_b^T = \{\bm{s}_i\}_{i=1} ^ {|\mathcal{D}_b^T|}$ for instances in $\mathcal{D}_b^T$ with $\mathcal{M}_{k-1}$ and $\mathcal{M}_{k}$.
		\STATE Sample a batch of actions $\mathcal{A}_b$ based on the probabilities estimated by $\pi(\bm{S}_b^T)$.
		\STATE Obtain the selected training batch $\mathcal{D}_{b'}^{T}$ according to $\mathcal{A}_b$.
		\STATE Update the target model with $\mathcal{D}_{b'}^{T}$ according to Equation \ref{eq:rl-loss}.
		\STATE Utilize the training loss of the current step and the previous step to obtain delayed reward $r$ according to Equation \ref{eq:reward}.
		\STATE Update policy model $\pi$ with $r$, $\bm{S}_b^T$ and $\mathcal{A}_b$ according to Equation \ref{eq:rl-opt}.
		\ENDFOR
	\end{algorithmic} 
\end{algorithm}

\section{An Empirical Study}\label{sec:exp}

In this section, we report a systematic empirical study using three well-accepted benchmark data sets and compare our proposed methods with a series of state-of-the-art methods.

\subsection{Datasets}
We use the datasets from CoNLL 2002 and 2003 NER shared tasks~\cite{DBLP:conf/conll/Sang02,DBLP:conf/conll/SangM03} with 4 distinct languages (Spanish, Dutch, English, and German). Moreover, to evaluate the generalization and scalability of our proposed framework, we select three non-western languages (Arabic, Hindi, and Chinese) from another multilingual NER dataset: WikiAnn~\cite{pan-etal-2017-cross}, partitioned by Rahimi et al.~\cite{rahimi-etal-2019-massively}.
Each datasets is split into training, development, and test sets. The statistics of those datasets are shown in Table~\ref{table:statistic}. All of those datasets are annotated with 4 types of entities, namely \emph{PER}, \emph{LOC}, \emph{ORG} and \emph{MISC} in \emph{BIO} tagging schema following~Wu et al.~\cite{DBLP:conf/acl/WuLKLH20} and Wu and Dredze~\cite{DBLP:conf/emnlp/WuD19}. The words are tokenized using WordPiece~\cite{wu2016google} and, following~Wu et al.~\cite{DBLP:conf/acl/WuLKLH20} and Wu and Dredze~\cite{DBLP:conf/emnlp/WuD19}, we only tag the first sub-word if a word is split.

For both CoNLL and WikiAnn, we use English as the source language and the others as target languages.
The pre-trained multilingual language models are fine-tuned using the annotated English training data. As for target languages, we remove the entity labels in the corresponding training data and adopt them as the unlabeled target-language instances. Note that to follow the zero-shot setting, we use the English development set to select the best checkpoints and evaluate them directly on the target language test sets.

\begin{table}[htbp]
\caption{\label{table:statistic} Statistics of the datasets.}
\vspace{-15pt}
\subtable{\bf \small (a) Statistics of CoNLL.}
{
\setlength{\abovecaptionskip}{3pt}
\setlength{\belowcaptionskip}{-3pt}
\small
\begin{center}
\begin{tabular}{l|cccc}
 \toprule
 \bf Language & \bf Type & \bf Train & \bf  Dev & \bf Test\\ 
\midrule
English-en  &   Sentence    &14,987     &3,466  & 3,684\\
(CoNLL-2003)&   Entity      &23,499     &5,942  & 5,648\\
\midrule
Spanish-es  &   Sentence    &8,323      &1,915  &1,517\\
(CoNLL-2002)&   Entity      &18,798     &4,351  &3,558\\
\midrule
Dutch-nl    &   Sentence    &15,806     &2,895  &5,195\\
(CoNLL-2002)&   Entity      &13,344     &2,616  &3,941\\
\midrule
German-de   &   Sentence    &12,705     &3,068  &3,160\\
(CoNLL-2003)&   Entity      &11,851     &4,833  &3,673\\
\bottomrule
\end{tabular}
\end{center}
}
\subtable{\bf \small (b) Statistics of WikiAnn.}
{
\setlength{\abovecaptionskip}{3pt}
\setlength{\belowcaptionskip}{-3pt}
\small
\begin{center}
\begin{tabular}{l|cccc}
 \toprule
 \bf Language & \bf Type & \bf Train & \bf  Dev & \bf Test\\ 
\midrule
English-en  &   Sentence    &20,000     &10,000  & 10,000\\
 &   Entity      &27,931     &14,146  &13,958\\
\midrule
Arabic-ar  &   Sentence    &20,000     &10,000  & 10,000\\
 &   Entity      &22,500     &11,266  &11,259\\
\midrule
Hindi-hi   &   Sentence    &5,000     &1,000  &1,000\\
 &   Entity      &6,124     &1,226  &1,228\\
\midrule
Chinese-zh  &   Sentence    &20,000     &10,000  & 10,000\\
 &   Entity      &25,031     &12,493  &12,532\\
\bottomrule
\end{tabular}
\end{center}
}
\end{table}

\subsection{Implementation Details}
We leverage the PyTorch version of cased multilingual \emph{BERT\textsubscript{base}} and \emph{XLM-R\textsubscript{base}} in \emph{HuggingFace's Transformers}\footnote{\url{https://github.com/huggingface/transformers}} as the basic encoders for all variants. Each of the two models has 12 Transformer layers, 12 self-attention heads and 768 hidden units (i.e.\ $d_h = 768$).
The hidden sizes of the policy network and the label embedding vector are set to $d_s = d_z = 256$ and $d_u = 50$, respectively.
We set the batch size to 64 and train each model for 5 epochs with a linear scheduler.
The parameters of embedding and the bottom three layers are fixed, following~Wu and Dredze~\cite{DBLP:conf/emnlp/WuD19}.
We use the AdamW optimizer~\cite{DBLP:journals/corr/abs-1711-05101} for all source and target models with a weight decay rate selected from \{5e-3, 7.5e-3\} and a learning rate chosen from \{3e-5, 5e-5, 7e-5\}.
The policy network is optimized using stochastic gradient descent and the learning rate is set to 0.01. The warm-up steps is selected from \{250, 500\}.
For evaluation, we use entity-level F1 score as the metric. 
The target models are evaluated on the development set every 100 steps and the checkpoints are saved based on the evaluation results.
Training of each round using 8 Tesla V100 GPUs takes 5 - 40 minutes depending on the encoder and target language.

\subsection{Baseline Models}
We compare our method with the following baselines.
    \cite{tackstrom-2012-nudging} is trained using cross-lingual word cluster features. \cite{DBLP:conf/conll/TsaiMR16,DBLP:conf/acl/NiDF17} leverage extra knowledge base or word alignment tools to annotate training data.
    \cite{DBLP:conf/emnlp/MayhewTR17,DBLP:conf/emnlp/XieYNSC18} generate target-language training data with machine translation. \cite{DBLP:conf/emnlp/WuD19,moon2019towards}  are trained using monolingual data and directly inference in the target language. \cite{DBLP:conf/aaai/WuLWCKHL20} leverages a meta-learning based method that benefits from similar instances. 
    \cite{DBLP:conf/acl/WuLKLH20} explores teacher-student paradigm to directly distill knowledge from a source language to a target language using the unlabeled data in the target language.
    \cite{DBLP:conf/emnlp/WuD20} proposes a contrastive alignment objective for multilingual encoders and outperforms previous word-level alignments.
    \cite{DBLP:conf/emnlp/PfeifferVGR20} introduces language, task, and invertible adapters to enable pre-trained multilingual models with high portability and efficient transfer capability.
We test statistical significance using t-test with p-value threshold $0.05$.

\subsection{Major Results}

Table~\ref{table:single-source} shows the results on cross-lingual NER of our method and the baseline methods. The first block compares our model (on top of mBERT) with the SOTA mBERT-based and the non-pretrained model based approaches. The second block compares our method (on top of XLM-R) with the SOTA XLM-R based models. The third block of Table~\ref{table:single-source} (a) denote works utilizing training data from multiple source languages. We can see that our proposed method significantly outperforms the baseline methods and achieves the new state-of-the-art performance.  The results clearly manifest the effectiveness of the proposed cross-lingual NER framework.

\begin{table}[t]
\caption{\label{table:single-source} The F1 scores of our method and the baseline models. Notes: $^{\dag}$ the reported results w.r.t.\ with the bottom three layers of language model fixed. $\P$ statistically significant improvements over Wu et al. ~\cite{DBLP:conf/acl/WuLKLH20}. $^{\ddag}$ the approaches utilizing training data from multiple source languages.}
\vspace{-15pt}
\subtable{\bf (a) Results on CoNLL.}
{
\resizebox{1.0\linewidth}{!}{
\begin{tabular}{l|ccc|c}
\toprule
\bf Method & \bf es & \bf nl & \bf de &\bf Average \\ 
\midrule
 
\citet{tackstrom-2012-nudging}                &59.30 &58.40 &40.40 &52.70\\

Tsai et al.~\cite{DBLP:conf/conll/TsaiMR16}              &60.55 &61.56 &48.12 &56.74\\

Ni et al.~\cite{DBLP:conf/acl/NiDF17}                  &65.10 &65.40 &58.50 &63.00\\

Mayhew et al.~\cite{DBLP:conf/emnlp/MayhewTR17}            &65.95 &66.50 &59.11 &63.85\\

Xie et al.~\cite{DBLP:conf/emnlp/XieYNSC18}              &72.37 &71.25 &57.76 &67.13\\

\citet{DBLP:conf/emnlp/WuD19}$^{\dag}$             &74.50 &79.50 &71.10 &75.03\\

\citet{moon2019towards}$^{\dag}$     &75.67 &80.38 &71.42 &75.82\\

Wu et al.~\cite{DBLP:conf/aaai/WuLWCKHL20}            &76.75 &80.44 &73.16 &76.78\\

Wu et al. ~\cite{DBLP:conf/acl/WuLKLH20} (mBERT)     &76.94 &80.89 &73.22 &77.02\\

\bf RIKD (mBERT)     &  \bf{77.84}$^{\P}$  & \bf{82.46}$^{\P}$ & \bf{75.48}$^{\P}$  & \bf{78.59}$^{\P}$ \\
\midrule
Wu et al. ~\cite{DBLP:conf/acl/WuLKLH20} (XLM-R)     &78.77 &80.99 &74.67 &78.14\\

\bf RIKD (XLM-R)     &  \bf{79.46}$^{\P}$  & \bf{81.40}$^{\P}$ & \bf{78.40}$^{\P}$  & \bf{79.75}$^{\P}$ \\

\midrule
\citet{tackstrom-2012-nudging}       $^{\ddag}$         &61.90 &59.90 &36.40 &52.73\\

\citet{moon2019towards}  $^{\ddag}$  &76.53 &83.35 &72.44 &77.44\\

Wu et al. ~\cite{DBLP:conf/acl/WuLKLH20}  $^{\ddag}$ &78.00 &81.33 &75.33 &78.22\\

\bottomrule
\end{tabular}
}
}

\subtable{\bf (b) Results on WikiAnn.}
{
\resizebox{1.0\linewidth}{!}{
\begin{tabular}{l|ccc|c}
\toprule
\bf Method & \bf ar & \bf hi & \bf zh &\bf Average \\ 
\midrule
\citet{DBLP:conf/emnlp/WuD20}     &42.30 &67.60 &52.90 &54.27\\
Wu et al. ~\cite{DBLP:conf/acl/WuLKLH20} (mBERT)     &43.12 &69.54 &48.12 &53.59\\

\bf RIKD (mBERT)     &  \bf{45.96}$^{\P}$  & \bf{70.28}$^{\P}$ & \bf{50.40}$^{\P}$  & \bf{55.55}$^{\P}$ \\

\midrule
Pfeiffer et al.~\cite{DBLP:conf/emnlp/PfeifferVGR20}     &41.80 & - &20.50 &-\\
\citet{DBLP:conf/emnlp/WuD20}     &45.50 &66.60 &43.90 &52.00\\
Wu et al. ~\cite{DBLP:conf/acl/WuLKLH20} (XLM-R)     &50.91 &72.48 &31.14 &51.51\\

\bf RIKD (XLM-R)     &  \bf{54.46}$^{\P}$  & \bf{74.42}$^{\P}$ & \bf{37.48}$^{\P}$  & \bf{55.45}$^{\P}$ \\
\bottomrule

\end{tabular}
}
}
\vspace{-5pt}
\end{table}

The pre-trained contextualized embedding based models \cite{moon2019towards,DBLP:conf/aaai/WuLWCKHL20,DBLP:conf/acl/WuLKLH20} outperform by a large margin those models learned from scratch.
Our mBERT and XLM-R based methods further achieve average gains of 1.57 and 1.61 percentage points over the strongest baseline (i.e., \cite{DBLP:conf/acl/WuLKLH20}) in F1 score on CoNLL, respectively. As for results of  non-western languages on WikiAnn, RIKD shows consistent improvements over \cite{DBLP:conf/acl/WuLKLH20} and non-distillation methods using additional layers~\cite{DBLP:conf/emnlp/PfeifferVGR20} and external resource~\cite{DBLP:conf/emnlp/WuD20}. (Note that Wu and Dredze~\cite{DBLP:conf/emnlp/WuD20} re-tokenize Chinese dataset and obtain relatively high results.) While the pre-trained methods directly adopt models trained on the source language, our proposed framework enables the model to take advantage of learning target language information from the unlabeled text and thus achieves even better performance.

In particular, \cite{DBLP:conf/acl/WuLKLH20} represents the previous state-of-the-art performance. It also employs a teacher-student framework to distill knowledge from a teacher model in the source language to the target model.
However, it directly transfers knowledge from a source model but neglects the noise in it.
Our proposed reinforced framework selectively transfers knowledge to reduce noise from the source language and fits the target languages better.

We further compare our method with the state-of-the-art multi-source cross-lingual NER approaches \cite{tackstrom-2012-nudging,moon2019towards,DBLP:conf/acl/WuLKLH20}, where human-annotated data on multiple source languages are assumed available. Although using only labeled data in English and unlabeled data in the target languages, our proposed method achieves the best average performance compared to the SOTA multi-source methods in our experiments. This further verifies that iterative knowledge distillation with reinforced instance selector is effective for the zero-shot cross-lingual NER task. 

From the application point of view, our method is more convenient and less data-consuming, especially for scenarios where multilingual training data is not available. Our proposed method also has great potentials in multi-source settings and other cross-lingual tasks, which are left for future work.

\subsection{Further Analysis}

\subsubsection{Ablation Study}

We conduct experiments on different variants of the proposed framework to investigate the contributions of different components.
Table~\ref{tab:ablation} presents the results of removing one component at a time.

\begin{table}[t]
\setlength{\abovecaptionskip}{3pt}
\setlength{\belowcaptionskip}{-3pt}
\centering
\small
\caption{F1 scores of ablation of reinforced selector and iterative knowledge distillation.  \emph{RIKD\textsubscript{Full}} is the proposed framework with iterative knowledge distillation and reinforced selectively transfer. \emph{$-$RL} removes Reinforced Knowledge Distillation from the full method. \emph{$-$IKD} removes Iterative Knowledge Distillation from the full method. \emph{$-$KD} removes knowledge distillation from the full method and directly conducts inference using the source model $\mathcal{M}_0$.}
\begin{tabular}{l | c c c |c } 
 \toprule
   &    \textbf{es} &    \textbf{nl} &   \textbf{de}   & \textbf{Average} \\
 \midrule
    \emph{RIKD\textsubscript{Full}} 
    &  79.46  & 81.40 & 78.40  & 79.75 \\
 \midrule
   \hspace{0.1cm} \emph{$-$RL    }  
    & 79.21 & 81.21 & 76.65 & 79.02 (0.73 $\downarrow$)\\
    \hspace{0.1cm} \emph{$-$IKD   }   
    & 78.90 & 81.02 & 74.86 & 78.26 (1.49 $\downarrow$)\\
    \hspace{0.1cm} \emph{$-$RL\&IKD }
    & 78.77 & 80.99 & 74.67 & 78.14 (1.60 $\downarrow$)\\
    \hspace{0.1cm} \emph{$-$KD }
    & 76.79 & 79.88 & 70.64 & 75.77 (3.97 $\downarrow$)\\
 \bottomrule
\end{tabular}
\label{tab:ablation}
\vspace{-5pt}
\end{table}

In general, all of the proposed techniques contribute to the cross-lingual setting. The full model consistently achieves the best performance on all languages experimented.
``\emph{$-$KD}'' is trained using the annotated data in the source language and directly infers in the target languages.  It suffers from a decrease of 3.97 points in F1 on average.
Both the single step and our iterative knowledge distillation settings outperform the direct model transfer approach, indicating that knowledge distillation is an efficient way to transfer knowledge across languages.
``\emph{$-$IKD}'' leads to a gain of 0.11 points in F1 on average compared to ``\emph{$-$RL\&IKD}'', which directly performs cross-lingual knowledge distillation.
Since the IKD step does not introduce extra supervision signals, we believe that the gain may come from that our iterative distillation framework benefits from better teacher models compared with the single round method.

The RL-based instance selection module contributes to both a single step and the iterative knowledge distillation framework.
Directly forcing the student model in the target language to imitate all behaviors of the teacher model may introduce source language specific bias~\cite{liu2020we} as well as the teacher model inference errors.
The intuition behind our instance selector is that selective distillation from the most informative cases can lead to better knowledge transfer performance on the target languages.  
The experimental results verify that the RL-based selector is capable of enhancing knowledge distillation by removing the noises in the teacher model prediction.

\subsubsection{Effect of Multiple Iterations in RIKD}
We further conduct experiments to study the effect of iterations in the RIKD training framework. Five rounds of iterations are conducted. For each iteration, the RL-based instance selector is leveraged to select instances for KD. The results are shown in Figure~\ref{fig:iterations}. Taking German as an example, RIKD gets gains over the last round by $4.22$, $1.98$, $1.56$ in the 1st, 2nd, and 3rd rounds, respectively. This further demonstrates the effectiveness of our proposed iterative training approach. It reduces noises in supervision signals step by step under the guidance of reinforcement learning. The figure also shows that the model performance stabilizes after three rounds. 
\begin{figure}[h]
\setlength{\abovecaptionskip}{3pt}
	\centering 
    \includegraphics[trim={0.5cm 0cm 0cm 1.9cm},clip,scale=0.22]{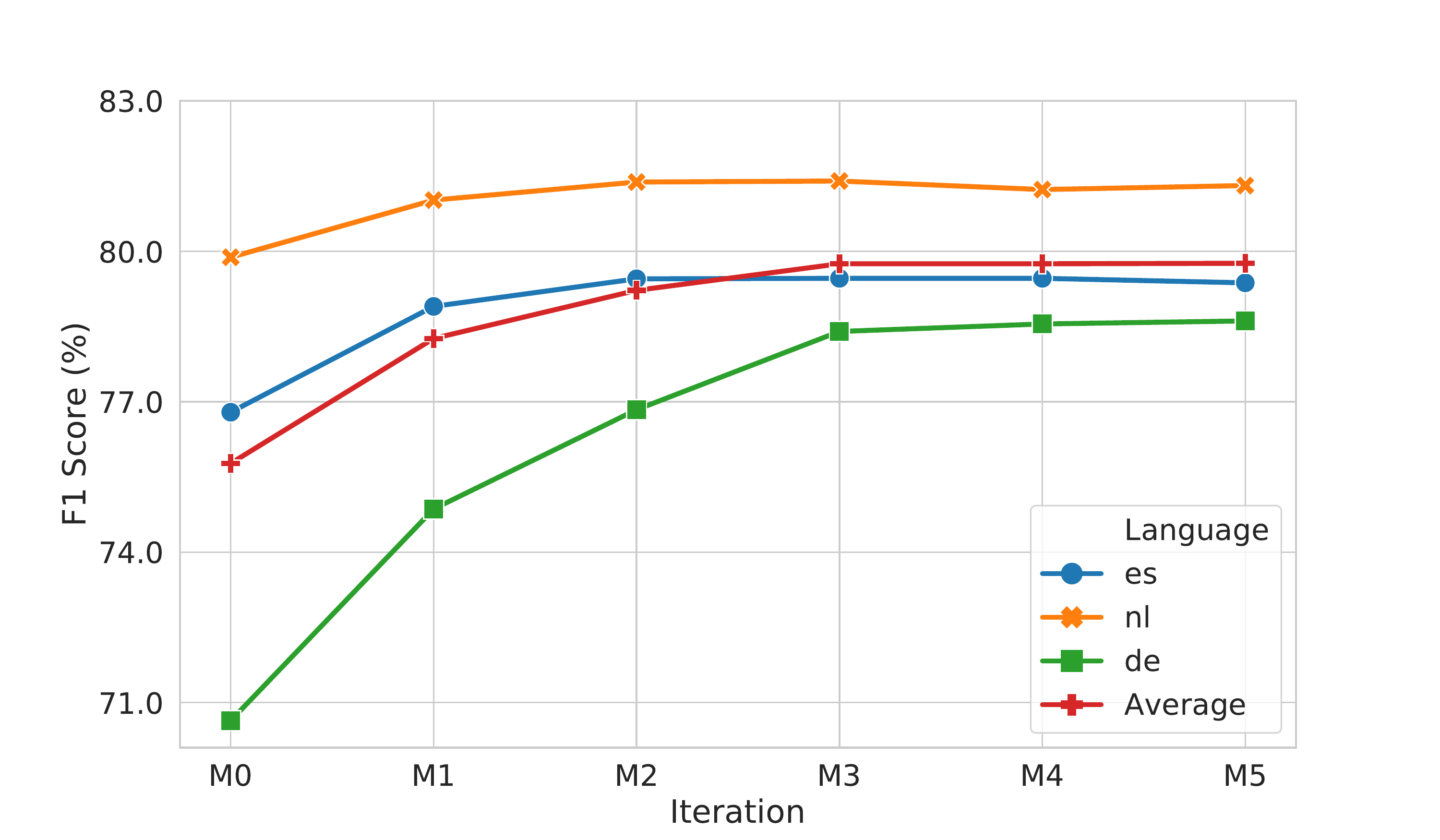}
	\caption{Effect of different iterations in RIKD (XLM-R) on CoNLL.} 
	\label{fig:iterations}
\vspace{-15pt}
\end{figure}

\subsubsection{Data Selection Strategy}
We study the effect of reinforced training instance selection by comparing with two other selection strategies including \emph{Confidence Selection} and \emph{Agreement Selection}. 

A straightforward method for training data selection is to keep those cases that the teacher model can predict with high confidence. Specifically, in an instance $\bm{x}$ of $L$ tokens, for each token, we use the entity label of the highest probability predicted by the teacher model as the entity label.  The confidence of $\bm{x}$ is the average token entity label probability.  That is,
    \begin{equation}
        conf(\bm{x}; \Theta_{teacher}) = \frac{1}{L}\sum_{i=1}^{L} \max_{\text{entity label }c} \{\tilde{\bm{p}}_c(x_i; \Theta_{teacher})\}
    \end{equation}
For each instance $\bm{x} \in \mathcal{D}_b^T$ with $conf(\bm{x}; \Theta_{teacher})$ over a predefined threshold $Thres_{conf}$, $\bm{x}$ is selected; otherwise, it is discarded. 

Another straightforward way for data selection is based on the agreement between the teacher model prediction and that of the student model. Specifically, the agreement score for each instance is defined as the minus of mean squared loss of the output probability distribution between the source and target model, that is,
    \begin{equation} \small
        agree(\bm{x}) = \\ 
        - \frac{1}{L}\sum_{i=1}^{L} \mathrm{MSE}(\tilde{\bm{p}}(x_i; \Theta_{teacher}), \; \tilde{\bm{p}}(x_i; \Theta_{student}))
    \end{equation}
For each instance $\bm{x} \in \mathcal{D}_b^T$, if $agree(\bm{x})$ passes a predefined threshold $Thres_{agree}$, $\bm{x}$ is selected otherwise discarded.

For fair comparisons, we set $Thres_{conf}$ and $Thres_{agree}$ properly to select a comparable amount of training instances in each batch as our RIKD method. As shown in Table~\ref{tab:ratio}, we record the number of discarded instances under our reinforced learning strategy and obtain the average discard ratio for each target language in iteration 1. Besides, we only adopt the performance of RIKD in iteration 1.

The results in Table~\ref{tab:strategy} show that the reinforced selection method achieves the best results, which verify that RIKD can select more informative cases for better knowledge transfer.

\begin{table}[t]
\setlength{\abovecaptionskip}{3pt}
\setlength{\belowcaptionskip}{-3pt}
\centering
\small
\caption{Average discard ratio of our approach. We first calculate the percentage of discarded cases in each batch and then average over all training batches in one iteration.}
\begin{tabular}{l | c c c c c c} 
 \toprule
   &    \textbf{es} &    \textbf{nl} &   \textbf{de} &  \textbf{ar} &    \textbf{hi} &    \textbf{zh}\\
 \midrule
    Discard ratio (\%) 
    & 38.8 & 56.7 & 52.4 & 60.8 & 23.5 & 57.2\\
 \bottomrule
\end{tabular}
\label{tab:ratio}
\vspace{-5pt}
\end{table}

\begin{table}[t]
\setlength{\abovecaptionskip}{3pt}
\setlength{\belowcaptionskip}{-3pt}
\centering
\caption{Analysis of different selection strategies.}
\resizebox{1.0\linewidth}{!}{
\begin{tabular}{l | c c c c c c |c } 
 \toprule
   &    \textbf{es} &    \textbf{nl} &   \textbf{de} &    \textbf{ar} &    \textbf{hi} &    \textbf{zh}   & \textbf{Average} \\
 \midrule
    \emph{RIKD\textsubscript{$\mathcal{M}_1$}} 
    & 78.90 & 81.02 & 74.86 & 52.38 & 72.88 & 32.14 & 65.36 \\
    \emph{Agreement Selection}  
    & 78.47 & 77.64 & 72.18 & 51.71 & 72.02 & 32.86 & 64.15 (1.21 $\downarrow$)\\
    \emph{Confidence Selection}   
    & 78.61 & 78.54 & 74.45 & 52.37 & 71.99 & 32.15 & 64.69 (0.67 $\downarrow$)\\
 \bottomrule
\end{tabular}
}
\label{tab:strategy}
\vspace{-5pt}
\end{table}

\subsubsection{State Vector Study}
This section investigates the effect of different features of the proposed state vector. Table~\ref{tab:state} shows the results when we remove each feature vector from the state vector introduced in Section~\ref{sec:RKD}. The results show that ablation of features causes performance to degrade to different extents. Among them $f_{conf}$, $f_{agree}$ and $f_{semantic}$ show the biggest contributions. One explanation is that these features could represent both the semantic meaning of the input training case but also how noisy the input training instance (in the target language) is. To be specific, $f_{semantic}$ encodes both the internal representation and the predicted labels of the training instance. $f_{conf}$ denotes the prediction confidence of the source model. $f_{agree}$ denotes the agreement degree (through MSE metric) between the predictions of source and target models. If the agreement is low, the training instance may have noise.  

\begin{table}[t]
\setlength{\abovecaptionskip}{3pt}
\setlength{\belowcaptionskip}{-3pt}
\centering
\caption{Analysis of different features in the state vector from Section~\ref{sec:RKD}.}
\small
\begin{tabular}{l | c c c |c } 
 \toprule
   &    \textbf{ar} &    \textbf{hi} &   \textbf{zh}   & \textbf{Average} \\
 \midrule
    \emph{RIKD\textsubscript{$\mathcal{M}_1$}} 
    & 52.38 & 72.88 & 32.14 & 52.47 \\
    - $f_{entities}$ 
    & 49.98 & 71.95 & 30.92 & 50.95 (2.52 $\downarrow$)\\
    - $f_{conf}$
    & 50.08 & 71.78 & 30.31 & 50.72 (1.75 $\downarrow$)\\
    - $f_{agree}$
    & 51.24 & 71.77 & 30.49 & 51.17 (1.30 $\downarrow$)\\
    - $f_{semantic}$
    & 49.20 & 71.68 & 28.05 & 49.64 (2.83 $\downarrow$)\\
    - $f_{length}$
    & 48.14 & 72.37 & 29.77 & 50.09 (2.38 $\downarrow$)\\
 \bottomrule
\end{tabular}
\label{tab:state}
\vspace{-5pt}
\end{table}

\subsubsection{Application to Industry Scenarios}

We further apply RIKD to one production scenario from the Microsoft Bing search engine to illustrate its practical effectiveness. The dataset consists of a human-annotated 50k en training set and multilingual test sets. Large scale unlabeled sentences are collected from web documents for distillation which includes 70.9k, 30.1k, 55.3k for es, hi, and zh respectively and we use the corresponding 5.6k, 3.2k, and 11k test data from the dataset for evaluation. 
As shown in Table~\ref{tab:industry}, RIKD outperforms Wu et al. ~\cite{DBLP:conf/acl/WuLKLH20} by 3.07 average F1 score. This manifests the effectiveness and generalization of our proposed method. 

\begin{table}[h]
\setlength{\abovecaptionskip}{3pt}
\setlength{\belowcaptionskip}{-3pt}
\centering
\caption{Results on industry scenario.}
\small
\begin{tabular}{l | c c c |c } 
 \toprule
   &    \textbf{es} &  \textbf{hi} &    \textbf{zh}   & \textbf{Average} \\
 \midrule
    Wu et al. ~\cite{DBLP:conf/acl/WuLKLH20}  
    & 77.94 & 65.61 & 31.17 & 58.24\\
    \bf RIKD
    & \textbf{79.59} & \textbf{68.29} & \textbf{36.06} & \textbf{61.31} (\textbf{3.07} $\uparrow$) \\
 \bottomrule
\end{tabular}
\label{tab:industry}
\vspace{-10pt}
\end{table}

\section{Conclusion and Industry Impact}\label{sec:con}
In this paper, we propose a reinforced knowledge distillation framework for cross-lingual named entity recognition (NER). The proposed method iteratively distills transferable knowledge from the model in the source language and performs language adaption using only unlabeled data in the target language.
We further introduce a reinforced instance selector that helps to selectively transfer useful knowledge from a teacher model to a student model.
We report a series of experiments on several widely used benchmark datasets. The results verify that the proposed framework outperforms the existing methods and establishes the new state-of-the-art performance on the cross-lingual NER task.

Moreover, RIKD is on the way to be deployed as an underlying technique in the Microsoft Bing search engine to serve many core modules such as Web ranking, Entity Pane, Answers Triggering, etc. And our framework will be adopted to improve User Intent Recognition and Slot Filling in a commercial voice assistant.

\begin{acks}
Shining Liang's research is supported by the National Natural Science Foundation of China (61976103, 61872161), the Scientific and Technological Development Program of Jilin Province (20190302029GX, 20180101330JC, 20180101328JC), and the Development and Reform Commission Program of Jilin Province (2019C053-8). Jian Pei's research is supported in part by the NSERC Discovery Grant program. All opinions, findings, conclusions and recommendations in this paper are those of the authors and do not necessarily reflect the views of the funding agencies.
\end{acks}

\bibliographystyle{ACM-Reference-Format}
\bibliography{main-rikd}

\end{sloppy}
\end{document}